\documentclass[10pt,twocolumn,letterpaper]{article}

\usepackage[pagenumbers]{cvpr} 

\usepackage[dvipsnames]{xcolor}

\definecolor{cvprblue}{rgb}{0.21,0.49,0.74}
\usepackage[pagebackref,breaklinks,colorlinks,citecolor=cvprblue]{hyperref}
\usepackage{multirow}
\usepackage{algorithm}
\usepackage{algorithmic}

\usepackage{amsmath}
\usepackage{amssymb}
\usepackage{arydshln}

\def\paperID{7083} 
\def\confName{CVPR}
\def\confYear{2024}

\title{StreamFlow: Streamlined Multi-Frame Optical Flow Estimation for Video Sequences}

\author{
{Shangkun Sun}~$^{1,2}$
~~~{Jiaming Liu}~$^{3}$
~~~{Thomas H. Li}~$^{1}$
~~~{Huaxia Li}~$^{3}$
~~~{Guoqing Liu}~$^{4}$
~~~{Wei Gao}~$^{1}$ \\
  $^1$School of Electronic and Computer Engineering, Peking University \\
  $^2$ Peng Cheng Laboratory, $^3$ Xiaohongshu Inc., $^4$ Minieye Inc. \\
}

\begin{document}
\maketitle
\begin{abstract}
Occlusions between consecutive frames have long posed a significant challenge in optical flow estimation. The inherent ambiguity introduced by occlusions directly violates the brightness constancy constraint and considerably hinders pixel-to-pixel matching. To address this issue, multi-frame optical flow methods leverage adjacent frames to mitigate the local ambiguity. Nevertheless, prior multi-frame methods predominantly adopt recursive flow estimation, resulting in a considerable computational overlap. In contrast, we propose a streamlined in-batch framework that eliminates the need for extensive redundant recursive computations while concurrently developing effective spatio-temporal modeling approaches under in-batch estimation constraints. Specifically, we present a Streamlined In-batch Multi-frame (SIM) pipeline tailored to video input, attaining a similar level of time efficiency to two-frame networks. Furthermore, we introduce an efficient Integrative Spatio-temporal Coherence (ISC) modeling method for effective spatio-temporal modeling during the encoding phase, which introduces no additional parameter overhead. Additionally, we devise a Global Temporal Regressor (GTR) that effectively explores temporal relations during decoding. Benefiting from the efficient SIM pipeline and effective modules, StreamFlow not only excels in terms of performance on the challenging KITTI and Sintel datasets, with particular improvement in occluded areas but also attains a remarkable $63.82\%$ enhancement in speed compared with previous multi-frame methods. \hyperlink{https://github.com/littlespray/StreamFlow}{Code} will be available soon.

\end{abstract}
\section{Introduction}
\label{sec:intro}





Optical flow estimation, which aims to model the per-pixel correspondence between two consecutive frames, is a fundamental task in computer vision. It has various downstream applications, such as video compression~\cite{dvc,dcvc}, object tracking~\cite{kale2015moving,choi2022moving}, and autonomous driving~\cite{capito2020optical,shi2022csflow}. 
Despite significant advancements in optical flow estimation in recent years, occlusion remains an issue that has not been fully resolved.
In particular, we consider occlusion as the disappearance of pixels in the current frame in the next frame~\cite{gma}, which violates the brightness consistency constraint and leads to great local ambiguity, significantly disrupting per-pixel matching.

\begin{figure}[htbp]
  \centering
  \includegraphics[width=\linewidth]{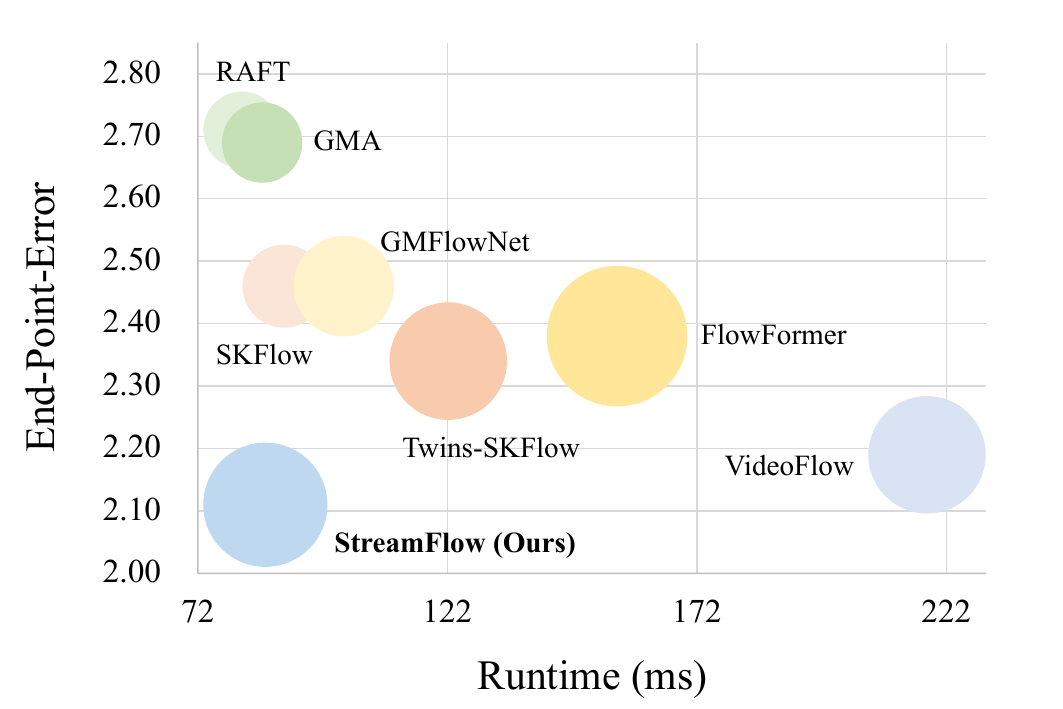}
  \caption{Comparison between performance, runtime, and parameters. A larger bubble represents more parameters. Models are trained via (C+)T schedule and validated on the Sintel final pass.}
  \label{fig:efficiency}
\end{figure}

\begin{figure}
  \centering
  \includegraphics[width=\linewidth]{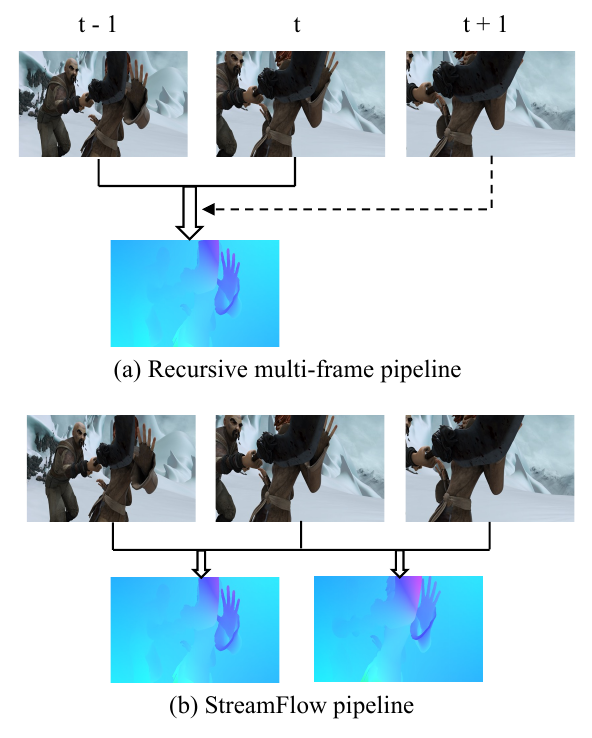}
  \caption{Comparison between different pipelines. Recursive methods leverage multi-frame for estimating two-frame flow, entailing substantial redundancy, while StreamFlow estimates multi-frame flows in-batch and eliminates overlapping computation.}
  \label{fig:pipeline}
\end{figure}

To alleviate this issue, prior research~\cite{raft,gma,gmflow,skflow,gmflownet,flowformer} has proposed various approaches based on a two-frame setup. More recently, there has been a growing interest in exploring temporal cues across multiple frames~\cite{transflow, videoflow, mfcflow, mfrflow}. Multi-frame optical flow methods utilize information from preceding and subsequent frames to better describe the temporal continuity of pixel motion, leading to a more accurate estimation of occluded motion. Nonetheless, when dealing with video inputs, previous multi-frame flow frameworks suffer from a considerable degree of redundant computation overlap, resulting in suboptimal efficiency, as exemplified in ~\cref{fig:pipeline}. For instance, TransFlow~\cite{transflow} devises a pure transformer architecture based on cross-frame attention and leverages self-supervised pre-training to better optimize the spatio-temporal modules. However, the computation of cross-frame attention still remains pairwise overlapping, and the pure transformer scheme is not advantageous in real-time applications. On the other hand, VideoFlow~\cite{videoflow} additionally predicts bidirectional flows and wins a remarkable performance gain. It successfully avoids redundant pairwise computations for bidirectional flows but still necessitates recursive estimation when predicting multiple unidirectional flows.

This gives rise to a core question: \emph{Is it possible to design a multi-frame pipeline that mitigates overlapping computations for video sequences while still effectively exploiting temporal cues and maintaining high efficiency in training and inference?}

In this work, we propose StreamFlow, a streamlined multi-frame optical flow estimation method tailored for video inputs. StreamFlow is made efficient through the Streamlined In-batch Multi-frame (SIM) pipeline, which avoids repetitive, overlapping computations when predicting unidirectional flows for video sequences. Furthermore, StreamFlow also explores the challenge of effectively modeling spatio-temporal cues under the constraint of non-overlapping in-batch estimation.
StreamFlow proposes a parameter-efficient Integrative Spatio-temporal Coherence (ISC) modeling module during encoding, and a Global Temporal Regressor (GTR) to decode all flows. Notably, these modules are quite lightweight, and StreamFlow attains comparable efficiency compared to two-frame methods with remarkable accuracy, as illustrated in~\cref{fig:efficiency}. Without self-supervised pre-training and the aim of bidirectional flows, StreamFlow achieves superior performance on Sintel and KITTI datasets, especially on the occluded regions.

In summary, our contributions are as follows:
\begin{itemize}
    \item We propose a Streamlined In-batch Multi-frame (SIM) pipeline for optical flow estimation, which eliminates the repetitive overlapping computation when computing unidirectional flows for video inputs.
    \item Under the constraint of a non-overlapping pipeline, we specifically designed the Integrative Spatio-temporal Coherence (ISC) module, which introduces no additional parameters and effectively exploits spatio-temporal cues.    
    \item For the SIM pipeline, we devise a Global Temporal Regressor (GTR) during decoding to further exploit temporal cues with modest additional computation cost.
    \item The proposed StreamFlow achieves superior performances on multiple benchmarks, particularly in occluded regions with comparable efficiency compared with two-frame methods, resulting in substantial improvements in optical flow estimation.
\end{itemize}

\section{Related work}
\label{sec:related_work}
\paragraph{Two-frame optical flow.}
Optical flow estimation in the form of a supervised learning task has been performed by FlowNet~\cite{flownet} using Convolutional Neural Networks (CNN). The encoder-decoder architecture of FlowNet predicts flow from coarse-to-fine using the hierarchy of the flow pyramid. Thereafter, a number of refined coarse-to-fine approaches~\cite{irr,pwcnet, pwcnet+,liteflownet1,liteflownet2,vcn,maskflownet,flownet2} emerged. The flow pyramid is constructed for the coarse-to-fine approach, which predicts the flow based on the flow guidance at a higher pyramid level. 
However, the flow guidance is often too coarse to capture small motions delicately and creates errors in later estimation. RAFT~\cite{raft} recently introduced an iterative all-pairs flow transform technique, which enables the prediction of high-resolution flow and recurrent refinement of the residual flow estimation. RAFT positively addresses the challenges of small motions and has consequently received high interest and performance in the field, 
inspiring numerous follow-up works~\cite{gma,agflow,skflow,gmflownet,gmflow,gaflow}.

\paragraph{Occlusions handling.}
Occlusion poses a great challenge to optical flow networks. 
It directly violates the brightness consistency constraint, which supposes pixels between adjacent frames remain the same brightness during the motion. 
The ambiguity brought by occlusions seriously interferes with the per-pixel matching as two-frame networks heavily rely on local evidence. Previous two-frame works mainly resolve the occluded pixels via multi-scale searching~\cite{pwcnet} or non-local modeling~\cite{gma,skflow,gmflownet,gmflow,flowformer,flowformer++}. These methods resolve the absent information to a certain extent. Nevertheless, in situations with severe occlusions, it becomes difficult to make up for the lack of local evidence without temporal cues, and the performance of two-frame networks remains limited in such scenarios.

\paragraph{Multi-frame optical flow.}
Exploiting temporal cues in optical flow estimation is an effective way to recover the occluded motion. Previous works~\cite{flowfusion, splatflow, continualflow, starflow, mfrflow, mfcflow,videoflow, transflow} propose various approaches to fuse temporal cues, such as leveraging previously predicted motion feature, optical flow, or contextual information. For instance, ContinualFlow~\cite{continualflow} uses previous flow priors to estimate the current occlusion map. STaRFlow~\cite{starflow} pass extracted features from different in multiple scales, jointly with occlusion maps. 
\cite{raft} proposes a warm-start strategy to initialize the original flow with the past flow before prediction. 
MFCFlow~\cite{mfcflow} and MFR~\cite{mfrflow} propose to leverage previously estimated motion features during decoding via feed-forward CNNs and self-similarity modeling, respectively. Nevertheless, these methods obtain a recursive strategy when handling video sequences, which divide the input sequence into lots of overlapping groups and take huge repeated computations.
TransFlow~\cite{transflow} decodes all flows simultaneously and achieves impressive results. However, it needs self-supervised pre-training on the flow datasets to help the temporal modeling modules converge. Besides, its pure transformer architecture and the overlapping computation when calculating cross-frame attention do not have advantages in terms of time.
VideoFlow~\cite{videoflow} additionally predicts the bi-direction flow to help the uni-direction flow estimation and win remarkable performance gain. Nevertheless, it still follows the recursive method to predict multiple unidirectional flows with the cost of predicting bidirectional flows.
In contrast, StreamFlow is proposed to avoid redundant, overlapping computation for consecutive unidirectional flow predictions while exploring efficient and effective temporal modules design under such a pipeline.

\begin{figure*}[htb]
  \centering
  \includegraphics[width=\linewidth]{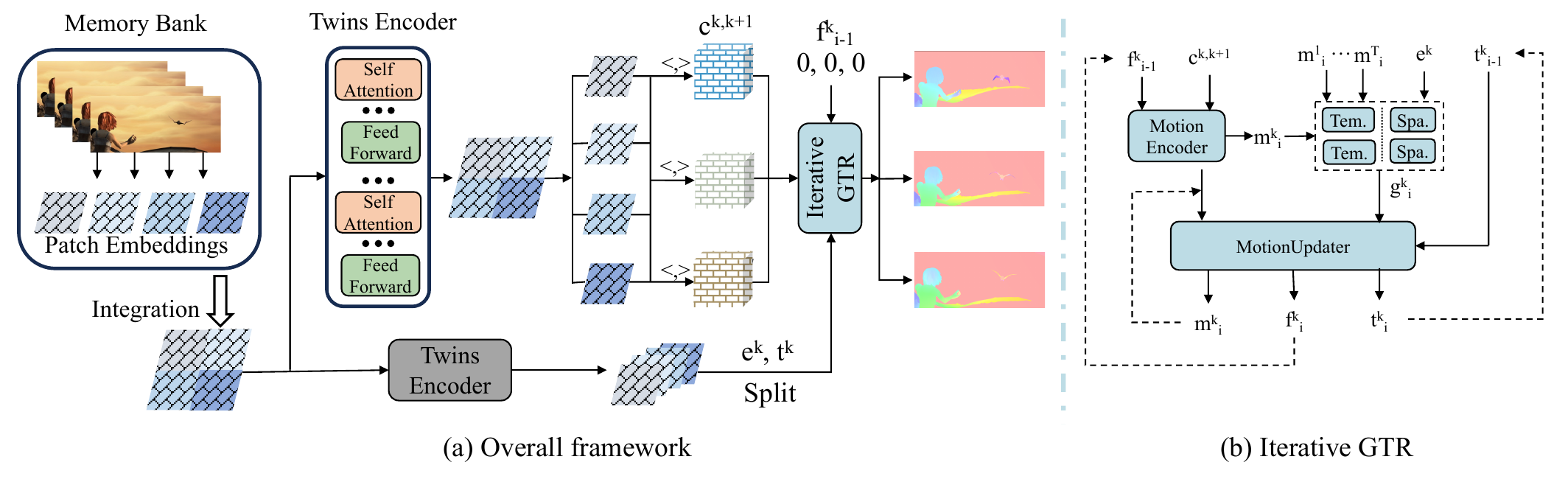}
  \caption{Overview of StreamFlow. (a) illustrates the overall framework and $<$,$>$ denotes the dot-product operation. (b) depicts the detailed module design of the GTR decoder.}
  \label{fig:overview}
\end{figure*}

\section{Methodology}
In this Section, we introduce StreamFlow, an efficient and effective in-batch framework for multi-frame optical flow estimation. The key components of StreamFlow consist of three parts: (1) The Streamlined In-batch Multi-frame (SIM) pipeline for efficient multi-frame estimation. (2) Integrative Spatio-temporal Coherence (ISC) modeling that is specifically designed for spatio-temporal modeling in the encoder of the SIM pipeline. (3) Global Temporal Regressor (GTR) that learns temporal relations for the SIM pipeline during decoding. We will first give an overview of our methods in ~\cref{sec:overview}, and then introduce each module in ~\cref{sec:sim}, ~\cref{sec:isc}, and ~\ref{sec:gtr}, respectively. In the end, we discuss the loss function design in ~\cref{sec:supervision}.

\subsection{Overview}
\label{sec:overview}
The overall framework of StreamFlow is illustrated in Figure~\ref{fig:overview}. For the basic encoder and decoder, similar to VideoFlow~\cite{videoflow}, StreamFlow adopts the Twins transformer~\cite{twins} as the encoder and utilizes the motion encoder and updater in SKFlow~\cite{skflow} during decoding. The overall iterative-refinement design that adopts an iterative decoder is the paradigm proposed in RAFT~\cite{raft} and followed by a lot of subsequent works~\cite{skflow, craft,gma,skflow,flowformer,flowformer++}. Input frames are first passed to two feature encoders that share the same architecture to extract the correlation feature and contextual feature, respectively. Then, the multi-scale all-pairs correlation vector is calculated based on the correlation feature. Namely, given feature embeddings $\mathbf{e}_{1}$ and $\mathbf{e}_{2}$ from the target frame and the reference frame, respectively:
\begin{equation}
\resizebox{1.0\linewidth}{!}{$\begin{aligned}
    \mathbf{c}^{l}(i, j, m, n) = 
    \frac{1}{2^{2l}} \sum\limits^{2^l}\limits_{u} \sum\limits^{2^l}\limits_{v} \left \langle \mathbf{e}_{1}(i, j), \mathbf{e}_{2}(2^{l}m+u, 2^{l}n+v) \right \rangle, 
\end{aligned}$}
\end{equation}
where the derived $\mathbf{c}^{l}(i, j, m, n)$ is the average over the correlation in the local $2^l \times 2^l$ window.
${l}$ denotes the $l th$ correlation level. $u$ and $v$ are the horizontal and vertical pixel motions, respectively. $\left \langle, \right \rangle$ refers to the dot product function. In summary, $\mathbf{c}^l(i, j, m, n)$ means the cost volume vector of $\mathbf{e}_{1}$ and $\mathbf{e}_{2}$ pooled with the $2^l \times 2^l$ kernel.

Then, the iterative decoder refines the flows via several updates. As depicted in ~\cref{fig:overview}, flows are initialized to zeros. The derived multi-scale correlation vector, extracted context feature, and the initialized flows are passed to the decoder, and then the refinement is conducted.


\subsection{Streamlined in-batch multi-frame pipeline}
\label{sec:sim}
As shown in~\cref{fig:pipeline}, previous multi-frame networks mainly compute recursively for the video inputs, resulting in a great deal of overlapping computation. Specifically, frames are divided into groups, and the flow between each frame in sequence is predicted recursively before processing the next group. The issue here lies in the overlap between frames within a group, where the same optical flow between overlapped frames would be calculated repeatedly. In contrast, StreamFlow is equipped with a Streamlined In-batch Multi-frame (SIM) Pipeline that tries to avoid redundancy. In the SIM pipeline, frames are divided into non-overlapping groups except for the initial frame. And in the same group, the repetitive computation is greatly reduced. First, each frame and its embeddings are stored in the memory bank so that the feature extraction and correlation construction are conducted only once. Besides, the spatio-temporal modeling methods are also designed specifically for non-overlapping computation, which will be given detailed discussion in ~\cref{sec:isc} and ~\cref{sec:gtr}. The pipeline is comparable to two-frame methods in latency with more accuracy and modest additional computation, as illustrated in ~\cref{fig:efficiency}.

\subsection{Integrative spatio-temporal coherence}
\label{sec:isc}
During the encoding process, we propose an Integrative Spatio-temporal Coherence (ISC) modeling method, especially for the SIM pipeline. Our design principles for temporal modeling modules in the decoder encompass two facets: firstly, adherence to the design criteria of the SIM pipeline, with a focus on minimizing pair-wise overlap operations, such as the computation of cross-frame attention between every pair of consecutive frames. Secondly, the modules should be efficient enough and not impede the overall speed of the network.

Therefore, we design the ISC method, which introduces no additional parameters and overlapping computation while learning spatio-temporal relations efficiently and effectively. The ISC method inherently takes the original modules in Twins. Specifically, after deriving patch embeddings from consecutive, ISC integrates temporally contiguous multiple input embeddings into a large feature embedding along the spatial dimension. Subsequently, it models the derived spatio-temporal graph using self-attention mechanisms and feed-forward layers in Twins, which could be formulated as,
\begin{align}
    x^{i}_{c} &= Integration^{T}_{t=1}(x^{j}_{t, c}), \\
    f(a_i, b_j) &= \frac{exp(a^T_i b_j /  \sqrt{d})}{ \sum^{N}_{j=1} exp(a^T_i b_j / \sqrt{d})} \\
    \mathbf{y}^{i}_{c} &= f(\mathbf{q}(\mathbf{x}^{i}_{c}), \mathbf{k}(\mathbf{x}^{i}_{c})) \mathbf{v}(\mathbf{x}^{i}_{c}), \\
    \mathbf{x}^{i}_{c} &= \mathbf{x}^{i}_{c} + \mathbf{W_{proj}}\mathbf{y}^{i}_{c}, \\
\end{align}
where $f(\cdot)$ is the attention function which conducts dot-product and softmax operation, $\mathbf{x}^{j}_{(t,c)}$ is the $j$th vector along spatial dimension at channel $c$ of the $t$th frame. $\mathbf{q}, \mathbf{k}$ and $\mathbf{v}$ is the derived query, key, and value vector $\mathbf{W_{proj}}$ is the projection matrix. By leveraging the derived spatio-temporal graph, the spatial and temporal relations are learned effectively, and no additional parameters are involved.

\subsection{Global temporal regressor}
\label{sec:gtr}
As for the decoder, we propose a Global Temporal Regressor (GTR) to predict and refine the predicted flows. Compared with the previous widely used decoder~\cite{raft, gma, skflow, agflow, sepflow, gaflow}, GTR introduces the temporal modeling module to exploit temporal cues from consecutive frames. Different from VideoFlow~\cite{videoflow} that concatenates motion features along a temporal dimension and implicitly learns temporal relations or TransFlow~\cite{transflow} that applies a transformer symmetric to the encoder, the core of GTR is super convolution kernels~\cite{skflow} and a lightweight temporal transformer block. The input correlation vectors, initialized flows, and contextual features are first passed into a motion encoder to derive motion features and then extracted for temporal and spatial features, which could be formulated as:
\begin{align}
    \mathbf{m}^{k}_{i} &= MotionEncoder(\mathbf{f}^{k}_{i-1}, \mathbf{c}^{k,k+1}), \\
    \mathbf{r}_{i} &= TemLayer^{T}_{j=1}(\mathbf{m}^{j}_{i}), \\
    \mathbf{s}^{k}_{i} &= SpaCrossAttn(\mathbf{m}^{j}_{i}, \mathbf{e}^{j}), \\
    \mathbf{g}^{k}_{i} &= Concat(\mathbf{r}_{i}, \mathbf{s}^{k}_{i}), \\
    \mathbf{t}^{k}_{i}, \mathbf{m}^{k}_{i}, \mathbf{\Delta{f}}^{k}_{i} &= MotionUpdater(\mathbf{m}^{k}_{i}, \mathbf{g}^{k}_{i}, \mathbf{t}^{k}_{i-1}), \\
    \mathbf{f}^{k}_{i} &= \mathbf{f}^{k}_{i-1} + \mathbf{\Delta{f}}^{k}_{i} 
\end{align}
where $\mathbf{m}^{k}_{i}$ is the derived motion feature of frame $k$ at the $i$th update and $\mathbf{f}^{k}_{i-1}$ denote the flow of frame $k$ after $i-1$th refinement. $\mathbf{c}_{k,k+1}$ denotes the correlation vector between frame $k$ and $k+1$. 
$MotionEncoder$ is the same motion encoder in the decoder of SKFlow~\cite{skflow}.
$\mathbf{r}_{i}$ denotes the temporal feature embedding extracted from the motion features of all frames. Notably, the caching mechanism of the MemoryBank is employed, thus necessitating the calculation of $\mathbf{r}_{i}$ only once for different frames.
$TempLayer$ is a lightweight temporal-learning layer that consists of temporal attention and feed-forward layers.
$\mathbf{e}^{k}$ refers to the feature embedding of frame $k$. Note that $e$ and $c$ are not updated during the refinement.
Inspired by the success of cross-attention mechanism in GMA~\cite{gma}, $SpaCrossAttn$ utilizes $\mathbf{m}^{k}_{i}$ and $\mathbf{e}^{k}$ to perform cross-attention. 
$\mathbf{t}$ denotes the extracted contextual information, which would be updated during each refinement.
In practice, the decoder estimates the residual of flow $\mathbf{\Delta{f}}^{k}_{i}$. And the final flow $\mathbf{f}^k$ is updated via $\mathbf{\Delta{f}}^{k}_{i}$ during each refinement.

\subsection{Supervision}
\label{sec:supervision}
StreamFlow adopts the overall loss in the same group as the total loss function. For each flow, StreamFlow adopts the same loss function as successful two-frame networks. Namely, the weighted sum for the predicted flows at different refinements. During both the training and the fine-tuning process, the supervision could be formulated as follows:
\begin{align}
    \mathcal{L} &= \sum_{k=1}^{T} \; \sum_{i=1}^{N} \; \theta^{N-i} \; || \mathbf{f}^k_{i} - \mathbf{f}^k_{gt} ||_1,
\end{align}
where $\mathbf{f}^k_{i}$ refers to the flow of frame $k$ at the $i$th refinement. $T$ and $N$ are the number of frames and refinements, respectively. $\theta$ denotes the weights on corresponding estimated flows. $\mathbf{f}_{gt}$ is the ground truth flow and $||\cdot||_1$ means the $l_1$ distance between ground truth and our predicted flow. In practice, $N$ is set to 12, $\theta$ is set to 0.8, the same as previous works~\cite{videoflow, raft, skflow, gma} for a fair comparison.

\section{Experiments}
\label{sec:experiments}

\begin{table*}[htb]
\centering
\scalebox{1.0}{
\begin{tabular}{llccccccc}
\toprule
\multirow{2}{*}{Training Data} & \multirow{2}{*}{Method} & \multicolumn{2}{c}{Sintel (train)} & \multicolumn{2}{c}{KITTI-15 (train)} & \multicolumn{2}{c}{Sintel (test)} & KITTI-15 (test) \\ \cmidrule(lr){3-4} \cmidrule(lr){5-6} \cmidrule(lr){7-8} \cmidrule(lr){9-9} 
&         & Clean           & Final           & Fl-epe           & Fl-all           & Clean           & Final          & Fl-all   \\ \midrule
\multirow{12}{*}{(C+)T}           & HD3~\cite{hd3}            & 3.84            & 8.77            & 13.17            & 24.0             & -           & -                 & -             \\
                                & VCN~\cite{vcn}            & 2.21            & 3.68            & 8.36            & 25.1             & -               & -              & -               \\
                                & FlowNet2~\cite{flownet2} & 2.02           & 3.54            & 10.08            & 30.0             & 3.96               & 6.02        & -          \\
                                & RAFT~\cite{raft}        & 1.43            & 2.71            & 5.04             & 17.4             & -               & -              & -     \\
                                & CRAFT~\cite{craft}        & 1.27            & 2.79           & 4.88           & 17.5             & -               & -              & -     \\
                                & GMA~\cite{gma}          & 1.30            & 2.74            & 4.69             & 17.1             & -               & -              & -          \\
                                & SKFlow~\cite{skflow}          & 1.22            & 2.46            & 4.27             & 15.5             & -               & -              & -         \\
                                & FlowFormer~\cite{flowformer}          & 1.00            & 2.45    & 4.09         & 14.7           & -     & -              & -  \\
                                & GAFlow~\cite{gaflow}                 & 1.02     & 2.45           & 3.98         & 15.0        & -        & -       & -        \\
                                & TransFlow~\cite{transflow}          & \underline{0.93}          & 2.33    & 3.98        & \underline{14.4}         & -               & -              & -  \\
                                & VideoFlow-BOF~\cite{videoflow}          & 1.03            & \underline{2.19}    & 3.96         & 15.3           & -               & -              & -  \\
                                & \textbf{Ours} & \textbf{0.87} & \textbf{2.11} & \textbf{3.85}  & \textbf{12.6} & - & - & - \\ \midrule
\multirow{18}{*}{(C+)T+S+K+H}    & LiteFlowNet2~\cite{liteflownet2}  & (1.30)          & (1.62)          & (1.47)           & (4.8)            & 3.48            & 4.69           & 7.74          \\
                              & IRR-PWC~\cite{irr}                & (1.92)          & (2.51)          & (1.63)           & (5.3)            & 3.84            & 4.58           & 7.65    \\
                              & MaskFlowNet~\cite{maskflownet}    & -               & -               & -                & -                & 2.52            & 4.17           & 6.10      \\
                              & Separable Flow\cite{sepflow}     & (0.69)          & (1.10)          & (0.69)           & (1.6)           & 1.50            & 2.67           & 4.64         \\
                              & PWC-Fusion~\cite{pwcnet+}                  & -          & -         & -          & -          & 3.43     & 4.57      & 7.17         \\
                              & StarFlow~\cite{starflow}                  & -          & -         & -          & -              & 2.72      & 3.71    & 7.65      \\
                              & RAFT$^\star$~\cite{raft}                  & (0.76)          & (1.22)          & (0.63)           & (1.5)           & 1.61      & 2.86      & 5.10       \\
                              & GMA$^\star$~\cite{gma}                    & (0.62)         & (1.06)           & (0.57)         & (1.2)         & 1.39        & 2.47        & 5.15        \\
                              & GMFlow~\cite{gmflow}                    & -        & -         & -       & -        & 1.74       & 2.90       & 9.32       \\
                              & GMFlowNet~\cite{gmflownet} & (0.59) & (0.91) & (0.64) & (1.5) & 1.39 & 2.65 & 4.79 \\
                              & AGFlow$^\star$~\cite{agflow}              & (0.65)          & (1.07)          & (0.58)           & (1.2)            & 1.43            & 2.47           & 4.89       \\
                              & SKFlow$^\star$~\cite{skflow}          & (0.52)        & (0.78)          & (0.51)         & (0.9)           & 1.28        & 2.27         & 4.84            \\
                              & FlowFormer~\cite{flowformer}          & (0.48)      & (0.74)        & (0.53)        & (1.1)          & 1.16            & 2.09          & 4.68          \\
                              & MFRFlow~\cite{mfrflow}                  & (0.64)          & (1.04)         & (0.54)          & (1.1)          & 1.55     & 2.80      & 5.03          \\
                              & MFCFlow~\cite{mfcflow}                  & (0.56)          & (0.89)         & (0.55)          & (1.1)          & 1.49    & 2.58         & 5.00   \\
                                & TransFlow~\cite{transflow}          & (0.42)      & (0.69)    & \underline{(0.49)}    & (1.05)   & 1.06  & 2.08  & \underline{4.32}  \\
                                & VideoFlow-BOF~\cite{videoflow}          & \underline{(0.37)}   & \underline{(0.54)}    & (0.52)   & \underline{(0.85)}    & \textbf{1.00}   & \textbf{1.71}  & 4.44  \\
                              & \textbf{Ours}           & \textbf{(0.28)}        & \textbf{0.38}          & \textbf{0.47}         & \textbf{0.77}           &  \underline{1.04}        & \underline{1.87}     & \textbf{4.24} \\
                              \bottomrule
\end{tabular}
}
\caption{Quantitative results on Sintel and KITTI. The average End-Point Error (EPE) is reported as the evaluation metric if not specified. $^\star$ refers to the warm-start strategy~\cite{raft} that use the previous flow for initialization. Bold and underlined metrics denote the method that ranks 1st and 2nd, respectively. Our method achieves superior performance on different benchmarks.}
\label{tab:1}
\end{table*}

\begin{figure}[htbp]
  \centering
  \includegraphics[width=\linewidth]{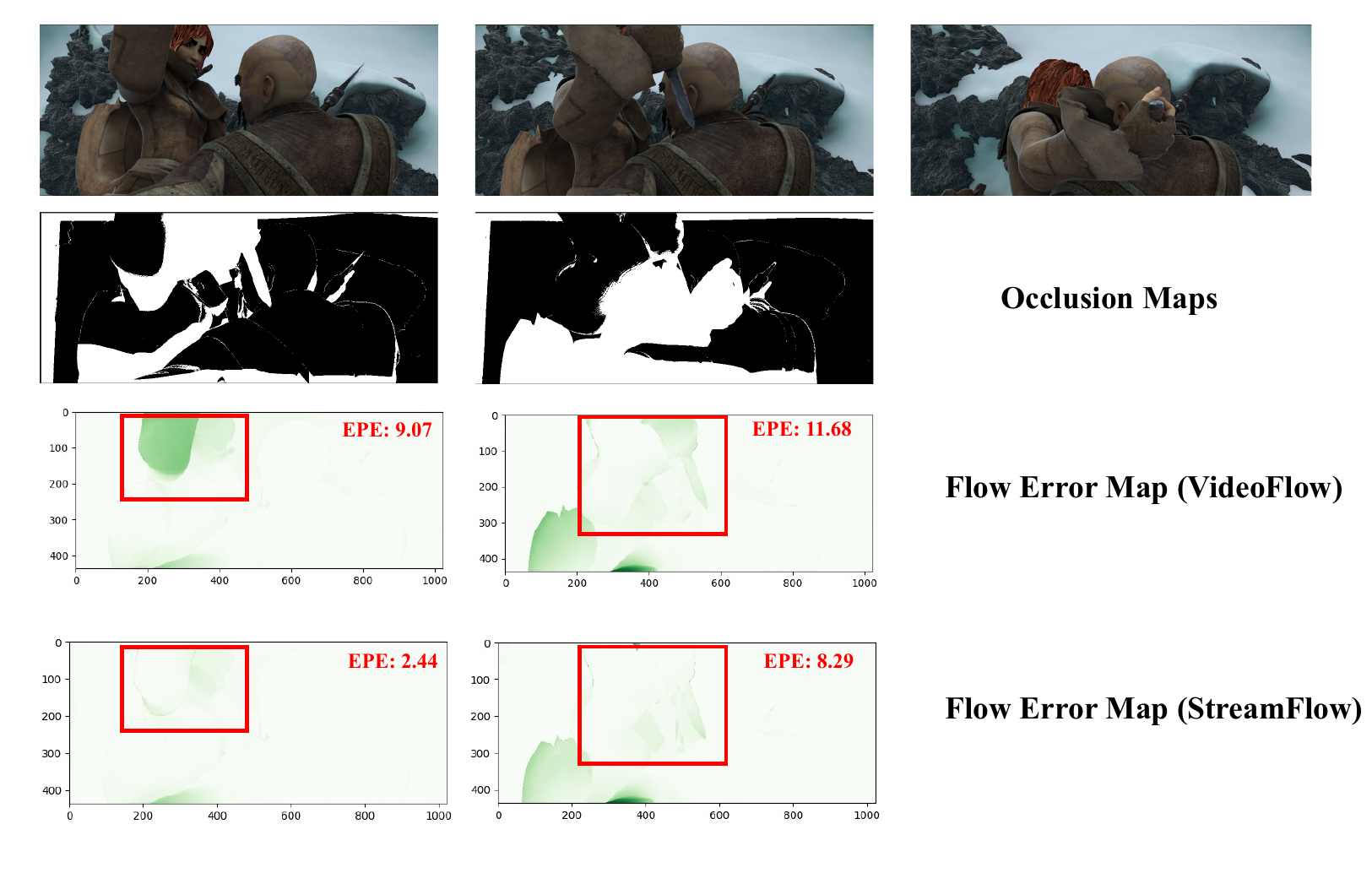} 
  \caption{Visualizations of the performance on the occluded regions. StreamFlow achieves comparable performance even with advanced methods. All models are trained on the FlyingThings dataset. A darker color in the flow error map denotes a higher estimation error compared with ground truth.}
  \label{fig:occ}
\end{figure}

\paragraph{Experimental setup.}
In this study, we evaluate our StreamFlow model on the Sintel~\cite{sintel} and KITTI~\cite{kitti15} datasets, following previous works~\cite{skflow,flowformer,raft}. 
In previous works, models are initially pre-trained on the FlyingChairs~\cite{flownet} and FlyingThings~\cite{flyingthings} datasets using the “C+T” schedule and then are subsequently fine-tuned using the “C+T+S+K+H” schedule on Sintel and KITTI datasets. In specific, for Sintel, models are fine-tuned on a combination of FlyingThings, Sintel, KITTI, and HD1K~\cite{hd1k}. After fine-tuning on Sintel, models are further fine-tuned using the KITTI dataset for the evaluation of KITTI.

\paragraph{Implementation details.}
Our StreamFlow method is built with PyTorch~\cite{pytorch} library, and our experiments are conducted on the NVIDIA A100 GPUs. During training, we adopt the AdamW~\cite{adamw} optimizer and the one-cycle learning rate policy~\cite{onecycle}, following previous works~\cite{raft,gma,skflow}.
During training, the number of refinements in the decoder is set to 12, following previous works.
Given the absence of multi-frame data information in the Chairs dataset, we follow VideoFlow~\cite{videoflow} to directly train on the FlyingThings dataset in the first stage.
The remaining training configurations remain consistent with prior works~\cite{videoflow,skflow,gma, raft}. The temporal and non-temporal modeling modules are concurrently trained.

\subsection{Quantitative Results}
From Table~\ref{tab:1}, we can learn that StreamFlow achieves superior performance on Sintel and KITTI. After being pre-trained on the FlyingThings dataset, StreamFlow demonstrates strong generalization ability across datasets. Given the leading performance of previous methods, StreamFlow could further reduce the end-point error by $0.16$ and $0.08$ on the challenging Sintel clean and final pass, respectively. On KITTI, StreamFlow outperforms the previous state-of-the-art method with $0.11$ and $17.65\%$ lower EPE and Fl-all metric.
Notably, without self-supervised pre-training or bi-directional flows, StreamFlow attains remarkable accuracy and efficiency on the challenging Sintel and KITTI benchmarks after the (C)+T and the +S+K+H schedule.


\begin{table*}[htb]
\centering
\scalebox{0.95}{
\begin{tabular}{llcccccccc}
\toprule
\multirow{2}{*}{Experiment}  & \multirow{2}{*}{Method} & \multicolumn{4}{c}{Sintel} & \multicolumn{2}{c}{KITTI} & Param & Latency \\ \cmidrule(lr){3-6} \cmidrule(lr){7-8} 
&      & Clean & Final & Occ (Albedo) & Noc (Albedo) & Fl-epe & Fl-all & (M) & (ms) \\ \midrule
\multirow{2}{*}{SIM pipeline}
& w/o & 1.03 & 2.34  & 7.69 & 0.35 & 4.64 & 14.70 & 12.49 & 122.18 \\ 
& \underline{w/} & 1.03 & 2.34  & 7.69 & 0.35 & 4.64 & 14.70 & 12.49 & \textbf{84.59} \\ \midrule 
\multirow{5}{*}{Temporal modules}
& w/o & 1.03 & 2.34  & 7.69 & 0.35 & 4.64 & 14.70 & \textbf{12.49} & \textbf{84.59} \\ 
& Temporal attn & \textbf{0.96} & 2.31  & 7.38 & 0.35 & 4.38 & 14.96 & 14.14 & 91.17 \\ 
& Pseudo 3D conv & 1.05 & 2.36  & 7.60 & 0.38 & 4.46 & 15.20 & 13.48 & 87.41 \\ 
& 3D conv & 0.98 & 2.34  & 7.63 & 0.33 & 4.57 & 15.59 & 16.03 & 93.05 \\ 
& \underline{ISC} & 0.97 & \textbf{2.29} & \textbf{7.11} & \textbf{0.32} & \textbf{4.14} & \textbf{14.16} & \textbf{12.49} & 88.35 \\ \midrule 
\multirow{3}{*}{Additional params}
& w/o & 0.97 & 2.29  & 7.11 & 0.32 & 4.14 & 14.16 & \textbf{12.49} & \textbf{84.59} \\ 
& w/ & 0.98 & 2.24  & 7.33 & \textbf{0.31} & 4.15 & 13.94 & 13.77 & 89.29 \\ 
& \underline{Ours} & \textbf{0.93} & \textbf{2.15} & \textbf{7.06} & \textbf{0.31} & \textbf{3.92} & \textbf{12.36} & 13.77  & 89.76  \\ \midrule  
\multirow{2}{*}{GTR module}
& w/o            & 0.97  & 2.29  & 7.11 & 0.32 & 4.14 & 14.16 & \textbf{12.49}  & \textbf{88.35} \\ 
& \underline{w/} & 0.93 & \textbf{2.15} & \textbf{7.06} & \textbf{0.31} & \textbf{3.92} & \textbf{12.36} & 13.77  & 89.76 \\ \midrule 
\multirow{2}{*}{ISC module}
& w/o & 1.01 & 2.19  & 7.23 & 0.33 & 4.06 & 13.95 & 13.77 & \textbf{86.02} \\  
& \underline{w/} & \textbf{0.93} & \textbf{2.15} & \textbf{7.06} & \textbf{0.31} & \textbf{3.92} & \textbf{12.36} & 13.77  & 89.76 \\ \midrule 
\multirow{2}{*}{Number of frames}
& 3 & 0.93 & 2.15  & 7.06 & 0.31 & 3.92 & \textbf{12.36} & \textbf{13.77} & 89.76 \\
& \underline{4} & \textbf{0.87} & \textbf{2.11}  & \textbf{6.24} & 0.31 & \textbf{3.85} & 12.62 & 14.25 & \textbf{85.53} \\
\bottomrule
\end{tabular}
}
\caption{Ablations on our proposed design. 
All models are trained using the "C+T" schedule and validated on Sintel.
The number of refinements is 12 for all methods. 
The settings used in our final model are underlined.}
\label{tab:ablation}
\end{table*}

\begin{figure*}[htbp]
  \centering
  \includegraphics[width=\linewidth]{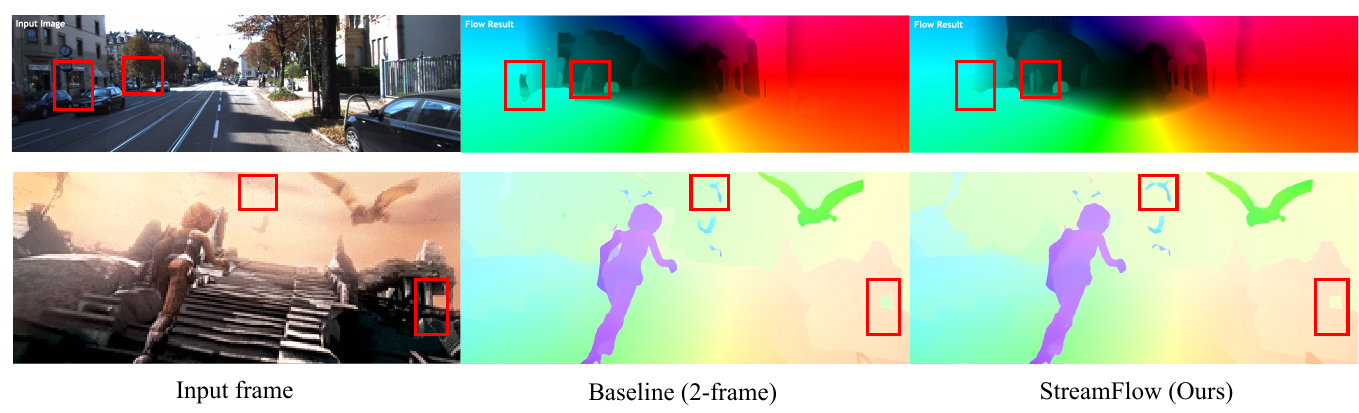}
  \caption{Visualizations of results on Sintel and KITTI test sets. Differences are highlighted with red bounding boxes. StreamFlow achieves fewer artifacts on both synthetic and real-world scenes.}
  \label{fig:vis}
\end{figure*}

\subsection{Occlusion Analysis}
In this section, we validate if StreamFlow could help improve the performance on the occlusions. We compare StreamFlow with its base two-frame model Twins-SKFlow, which strengthens SKFlow~\cite{skflow} with the Twins~\cite{twins} encoder. Evaluations are conducted on the matched and unmatched areas of the challenging Sintel test dataset. 
The matched areas denote regions visible in adjacent frames and the unmatched areas refer to regions visible only in one of two adjacent frames. Our models are trained using the T+S+H+K schedule. We could learn that StreamFlow attains remarkable improvements on occluded areas, as shown in~\cref{tab:occ}. We also visualize the performance on occluded regions in~\cref{fig:occ}. On the challenging Sintel final test set, StreamFlow attains the improvement of $10.77\%$ and $11.83\%$ on unmatched and matched regions, respectively. On the clean pass, StreamFlow improves the performance by $15.53\%$, $15.56\%$, and $15.45\%$ on unmatched, matched, and overall regions. We could learn that StreamFlow improves not only the flow estimation in unmatched regions but also the estimation in matched regions.

\begin{table}[htb]
\centering
\scalebox{0.85}{
\begin{tabular}{lcccccc}
\toprule
\multirow{2}{*}{Method}  & \multicolumn{3}{c}{Clean} & \multicolumn{3}{c}{Final} \\ \cmidrule(lr){2-4} \cmidrule(lr){5-7}
 & Unm. & Mat. & All & Unm. & Mat. & All \\ \midrule
GMFlow~\cite{gmflow} & 10.56 & 0.65 & 1.74 & 15.80 & 1.32 & 2.90 \\ \midrule
GMFlowNet~\cite{gmflownet} & 8.49 & 0.52 & 1.39 & 13.88 & 1.27 & 2.65 \\ \midrule
SKFlow~\cite{skflow} & 7.24 & 0.55 & 1.28 & 11.51 & 1.46  & 2.28 \\ \midrule
FlowFormer~\cite{flowformer++} & 7.16 & 0.42 & 1.16 & 11.30 & 0.96 & 2.09 \\ \midrule
TransFlow~\cite{transflow} & 6.77 & 0.36 & 1.06 & 10.96 & 0.99  & 2.08 \\ \midrule
Baseline & 7.60  & 0.45 & 1.23 & 11.70 & 0.93 & 2.11 \\ \midrule
Ours & \textbf{6.42} & \textbf{0.38} & \textbf{1.04} & \textbf{10.44} & \textbf{0.82} & \textbf{1.87} \\ 
\bottomrule
\end{tabular}
}
\caption{Occlusion analysis on Sintel test set. Unm. and Mat. denote performance on unmatched and matched areas, respectively.}
\label{tab:occ}
\end{table}

\subsection{Abaltions}
In this section, we verify the effectiveness of StreamFlow designs, as shown in~\cref{tab:ablation}. 
For a fair comparison, all models in the same experiment are trained under the same settings on the FlyingThings dataset. Then we evaluate each method on Sintel and KITTI. 
Below we will introduce each experiment in more detail. 

\paragraph{SIM pipeline.} We test the efficiency of the vanilla recursive pipeline and our SIM pipeline. Recursive methods utilize multi-frames to predict the flow of the current two frames and bring substantial redundant computation, while the SIM pipeline estimates multiple flows concurrently and minimizes the overlapping calculation. As shown in~\cref{tab:ablation}, the SIM pipeline brings great gain in efficiency.

\paragraph{Temporal modules.} In this part, we explore the performance and efficiency of different temporal modeling methods in the flow encoder. Temporal attn refers to applying a temporal attention layer after each spatial self-attention modeling in Twins. Pseudo conv~\cite{pseudo1dconv} denotes stacking 1D convolution layers in the temporal dimension to imitate 3D convolutions at minimal cost. We also apply 3D convolutions at the end of the flow encoder to learn temporal relations. As shown in~\cref{tab:ablation}, our ISC module achieves a good trade-off between efficiency and effectiveness. The improvements achieved by other methods are not as pronounced. We hypothesize that the limited volume of optical flow data impedes the efficient training of the spatio-temporal module from scratch to accomplish good optimization. For comparison, VideoFlow does not apply temporal modeling modules in the encoder, and TransFlow~\cite{transflow} applies self-supervised pre-training for better optimization.

\paragraph{Additional params.} In this part, we aim to determine whether the performance gain is due to the additional
parameters or the effective temporal modeling method. To
this end, we introduce the additional parameters by widening the baseline network. 
Namely, we extract higher-dimension features along the spatial dimension and concatenate them with the original motion feature.
All models in this section are equipped with the ISC module.
``w/o" denotes the baseline Twins-SKFlow network. ``w" means adding additional parameters.
``Ours" denotes the method equipped with our temporal modeling modules. Results show the improvement achieved by simply adding more parameters is
minor, and the performance gain is primarily attributed to
the effectiveness of StreamFlow modules.

\paragraph{GTR module.} We also examined whether the GTR module could enhance flow predictions. ``w/o" means applying vanilla SKFlow decoder while ``w" denotes using GTR. 
All models in this part utilize the ISC module in the encoder.
~\cref{tab:ablation} demonstrates the necessity of incorporating the GTR. With GTR, StreamFlow could further achieve stable improvement on multiple benchmarks. We could also learn that GTR especially helps the flow estimation on the challenging final passes, with the performance gain of $0.14$.

\paragraph{ISC module.} In this part, we verify the effectiveness of the proposed ISC module. All models in this part adopt GTR as the flow decoder. From~\cref{tab:ablation}, we could learn that the ISC module is efficient and effective in temporal modeling and makes a significant contribution to the improvement of the multiple-frame pipeline. It introduces no additional parameters and a modest increase in runtime, while significantly boosting the performance.

\paragraph{Number of frames.} We delve into the influence of different numbers of input frames, as illustrated in~\cref{tab:ablation}. We set the number of frames to 4 due to limitations in GPU memory. From an efficiency standpoint, augmenting the number of input frames results in a higher proportion of redundant computations eliminated by StreamFlow within the total computational workload, consequently leading to a more substantial improvement in processing time. Although there is an increase in the parameter count for temporal modeling, the efficiency of StreamFlow is further enhanced in the context of four input frames due to a reduced proportion of redundant computations, resulting in a shorter average prediction time per frame compared to the three-frame setting.

\subsection{Qualitative results}
In this section, we demonstrate visualization results on both synthetic and real-world scenes. We test the models on the challenging Sintel~\cite{sintel} and KITTI~\cite{kitti15}, as shown in~\cref{fig:vis}. In the appendix, we also demonstrate the qualitative performance on the real-world dataset DAVIS~\cite{davis17}. Our models are pre-trained using the T+H+S+K schedule. We could learn that StreamFlow could still achieve remarkable qualitative results when generalized to real-world scenes.

\subsection{Efficiency analysis}
In this section, we evaluate the efficiency of the StreamFlow
method in terms of runtime and parameter counts. Our experiments were conducted on an NVIDIA A100 GPU. Models are trained using the (C+)T schedule and evaluated on the Sintel dataset. The runtime is measured as the average inference time per frame of five runs on the Sintel training set. Figure~\cref{fig:efficiency} depicts the results, where the size of the bubble corresponds to the number of parameters, the horizontal axis represents time, and the vertical axis represents end-point-error. We could learn StreamFlow achieves nearly comparable efficiency with state-of-the-art two-frame methods while achieving superior performance. The key to maintaining high efficiency is its non-overlapping SIM pipeline. StreamFlow does not perform pairwise redundant computation and predicts all flows simultaneously. Another reason for the high speed is the CNN-based decoder of StreamFlow. We could learn that StreamFlow is much faster than the pure two-frame transformer architecture FlowFormer. Besides, the specially designed lightweight temporal-modeling modules also contribute to the efficiency, simultaneously aiding in better results compared to the 2-frame baseline Twins-SKFlow.


\section{Conclusion}
In this work, we proposed StreamFlow, a multi-frame
optical flow estimation approach proficient in identifying optical flow across multiple video frames using efficient Spatio-temporal relationship mining. 
StreamFlow proposes to estimate multi-frame optical flow via an in-batch method (SIM pipeline) and explores the design of temporal modeling modules under such constraints. In specific, StreamFlow introduces a parameter-efficient Integrative Spatio-temporal Coherence (ISC) module that is seamlessly equipped with the encoder, and designs an efficient and effective Global Temporal Regressor (GTR) module in the decoder. Extensive experiments demonstrate the efficiency and effectiveness of StreamFlow. 
With the proposed SIM pipline, ISC, and GTR module, StreamFlow showed comparable efficiency with two-frame methods while achieving remarkable accuracy, especially in occluded regions.
\clearpage
\setcounter{page}{1}
\maketitlesupplementary

\begin{figure*}[htbp]
  \centering
  \includegraphics[width=\linewidth]{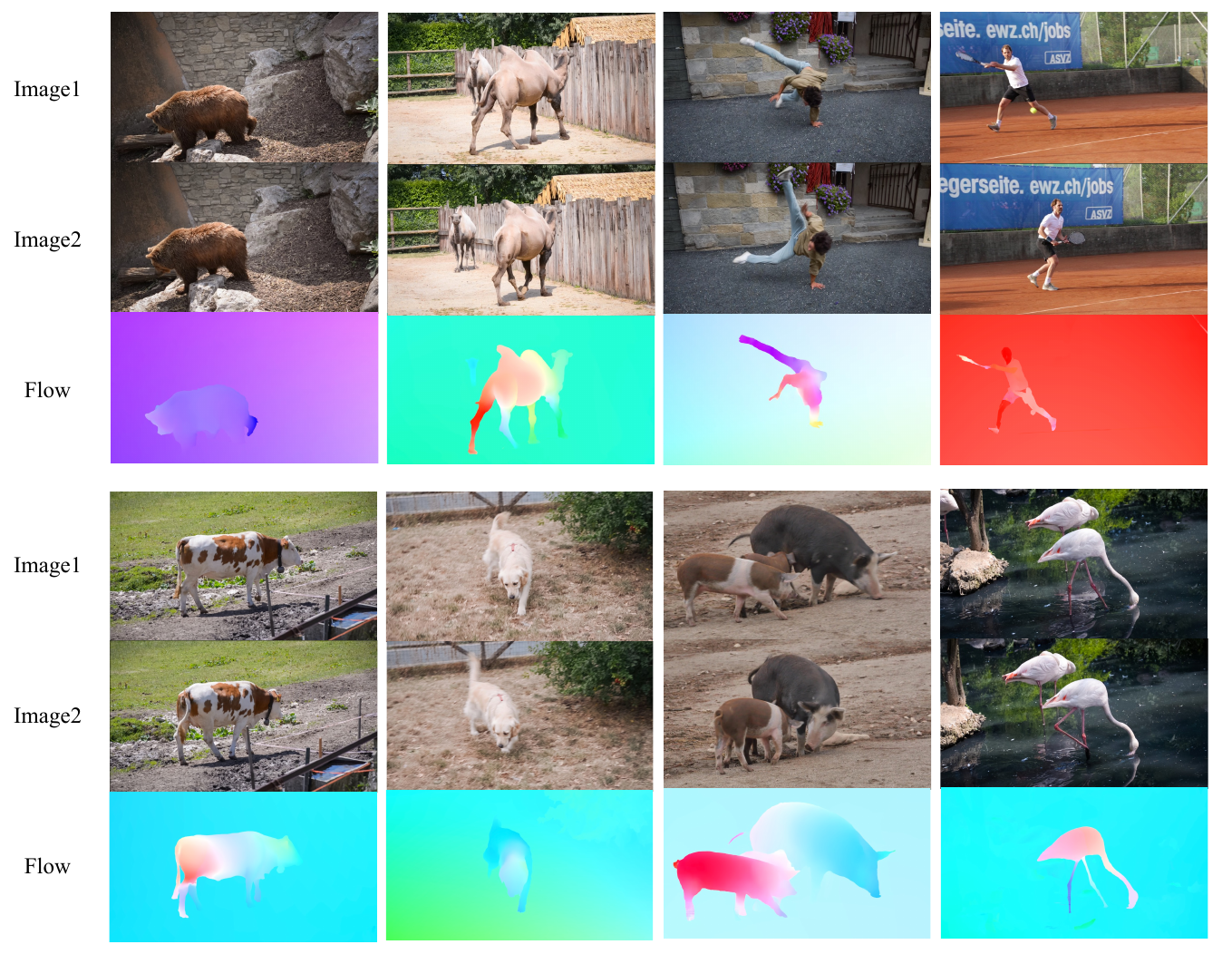}
  \caption{Visualizations of predicted flows on DAVIS~\cite{davis17}. StreamFlow demonstrates robust generalization to other real-world datasets, performing well in challenging scenarios for optical flow estimation, as evidenced by instances such as the occluded hind legs of the bear in the first column and the small tennis ball in the last column.}
  \label{fig:davis}
\end{figure*}

\paragraph{Qualitative analysis on real-world scenes}
\label{sec:davis}
In this section, we facilitate our visualizations and evaluations using two prominent real-world datasets, namely DAVIS~\cite{davis17}. The DAVIS dataset, short for Densely Annotated VIdeo Segmentation, is a widely recognized benchmark in the field of computer vision. It comprises high-quality video sequences captured in diverse scenarios, encompassing a broad range of challenging visual conditions such as occlusions, motion blur, and dynamic object interactions. The dataset provides pixel-level annotations for every frame, facilitating precise evaluation and comparison of various video segmentation methods. The visualizations on the DAVIS dataset is shown in ~\cref{fig:davis}. Our model is pretrained using the ``T" and ``T+S+H+K" schedule and then fine-tuned on KITTI~\cite{kitti15}. ``T" denotes the FlyingThings~\cite{flyingthings} dataset and ``T+S+H+K" refers to te combination of the FlyingThings, Sintel~\cite{sintel}, HD1K~\cite{hd1k}, and KITTI datasets. Then we infer our models on the DAVIS dataset. The number of refinements is set to 12. The number of input frames for each non-overlapping group is 3. We could learn that StreamFlow demonstrates remarkable adaptability across real-world datasets, showing its robust performance in challenging scenes for optical flow estimation. This is particularly evident in scenarios such as the occlusion of the bear's hind legs in the first row, first column and the small motion of the small tennis ball in the last column. Additionally, it can be observed that in the motion captured in the first row, second and third columns, the hind legs of the camel and the leg movements of the dancer are also vividly delineated. These instances reaffirm its efficacy in diverse and demanding environments for optical flow estimation.

\section{Initialization of GTR}
In this section, we investigate the impact of different GTR initialization methods. Previous works in spatio-temporal modeling such as~\cite{timesformer} have suggested initializing the temporal modules with zero values. We employed two distinct initialization approaches, namely zero initialization and PyTorch's default initialization, and the corresponding results are presented in~\cref{tab:init}. Following training on the FlyingThings dataset, the model was tested on the Sintel and KITTI datasets. It is evident from the results that the zero initialization could contributes to a better overall performance.

\begin{table*}[tb]
\centering
\begin{tabular}{lcccc}
\toprule
Method & Sintel (Clean) & Sintel (Final) & KITTI (EPE) & KITTI (Fl-all) \\ \midrule
Default & \textbf{0.91} & 2.20  & 4.05 & 13.44 \\  
Zero-init & 0.93 & \textbf{2.15} & \textbf{3.92} & \textbf{12.36} \\ 
\bottomrule
\end{tabular}
\caption{Comparison of different ways of initialization. All models are trained under the FlyingThings.}
\label{tab:init}
\end{table*}


{
    \small
    \bibliographystyle{ieeenat_fullname}
    \bibliography{main}

\begin{thebibliography}{46}
\providecommand{\natexlab}[1]{#1}
\providecommand{\url}[1]{\texttt{#1}}
\expandafter\ifx\csname urlstyle\endcsname\relax
  \providecommand{\doi}[1]{doi: #1}\else
  \providecommand{\doi}{doi: \begingroup \urlstyle{rm}\Url}\fi

\bibitem[Angelino et~al.(2010)Angelino, Yamins, and Seltzer]{starflow}
Elaine Angelino, Daniel Yamins, and Margo Seltzer.
\newblock Starflow: A script-centric data analysis environment.
\newblock In \emph{Provenance and Annotation of Data and Processes: Third International Provenance and Annotation Workshop, IPAW 2010, Troy, NY, USA, June 15-16, 2010. Revised Selected Papers 3}, pages 236--250. Springer, 2010.

\bibitem[Bertasius et~al.(2021)Bertasius, Wang, and Torresani]{timesformer}
Gedas Bertasius, Heng Wang, and Lorenzo Torresani.
\newblock Is space-time attention all you need for video understanding?
\newblock In \emph{ICML}, page~4, 2021.

\bibitem[Butler et~al.(2012)Butler, Wulff, Stanley, and Black]{sintel}
Daniel~J Butler, Jonas Wulff, Garrett~B Stanley, and Michael~J Black.
\newblock A naturalistic open source movie for optical flow evaluation.
\newblock In \emph{European conference on computer vision}, pages 611--625. Springer, 2012.

\bibitem[Capito et~al.(2020)Capito, Ozguner, and Redmill]{capito2020optical}
Linda Capito, Umit Ozguner, and Keith Redmill.
\newblock Optical flow based visual potential field for autonomous driving.
\newblock In \emph{2020 IEEE Intelligent Vehicles Symposium (IV)}, pages 885--891. IEEE, 2020.

\bibitem[Chen et~al.(2023)Chen, Zhu, Shi, Zhang, Zhang, Zhang, and Li]{mfcflow}
Yonghu Chen, Dongchen Zhu, Wenjun Shi, Guanghui Zhang, Tianyu Zhang, Xiaolin Zhang, and Jiamao Li.
\newblock Mfcflow: A motion feature compensated multi-frame recurrent network for optical flow estimation.
\newblock In \emph{Proceedings of the IEEE/CVF Winter Conference on Applications of Computer Vision}, pages 5068--5077, 2023.

\bibitem[Choi et~al.(2022)Choi, Kang, and Kim]{choi2022moving}
Hosik Choi, Byungmun Kang, and DaeEun Kim.
\newblock Moving object tracking based on sparse optical flow with moving window and target estimator.
\newblock \emph{Sensors}, 22\penalty0 (8):\penalty0 2878, 2022.

\bibitem[Chu et~al.(2021)Chu, Tian, Wang, Zhang, Ren, Wei, Xia, and Shen]{twins}
Xiangxiang Chu, Zhi Tian, Yuqing Wang, Bo Zhang, Haibing Ren, Xiaolin Wei, Huaxia Xia, and Chunhua Shen.
\newblock Twins: Revisiting the design of spatial attention in vision transformers.
\newblock \emph{Advances in Neural Information Processing Systems}, 34:\penalty0 9355--9366, 2021.

\bibitem[Dosovitskiy et~al.(2015)Dosovitskiy, Fischer, Ilg, Hausser, Hazirbas, Golkov, van~der Smagt, Cremers, and Brox]{flownet}
Alexey Dosovitskiy, Philipp Fischer, Eddy Ilg, Philip Hausser, Caner Hazirbas, Vladimir Golkov, Patrick van~der Smagt, Daniel Cremers, and Thomas Brox.
\newblock Flownet: Learning optical flow with convolutional networks.
\newblock In \emph{Proceedings of the IEEE International Conference on Computer Vision (ICCV)}, 2015.

\bibitem[Huang et~al.(2022)Huang, Shi, Zhang, Wang, Cheung, Qin, Dai, and Li]{flowformer}
Zhaoyang Huang, Xiaoyu Shi, Chao Zhang, Qiang Wang, Ka~Chun Cheung, Hongwei Qin, Jifeng Dai, and Hongsheng Li.
\newblock Flowformer: A transformer architecture for optical flow.
\newblock \emph{arXiv preprint arXiv:2203.16194}, 2022.

\bibitem[Hui et~al.(2018)Hui, Tang, and Loy]{liteflownet1}
Tak-Wai Hui, Xiaoou Tang, and Chen~Change Loy.
\newblock Liteflownet: A lightweight convolutional neural network for optical flow estimation.
\newblock In \emph{Proceedings of the IEEE conference on computer vision and pattern recognition}, pages 8981--8989, 2018.

\bibitem[Hui et~al.(2020)Hui, Tang, and Loy]{liteflownet2}
Tak-Wai Hui, Xiaoou Tang, and Chen~Change Loy.
\newblock A lightweight optical flow cnn—revisiting data fidelity and regularization.
\newblock \emph{IEEE transactions on pattern analysis and machine intelligence}, 43\penalty0 (8):\penalty0 2555--2569, 2020.

\bibitem[Hur and Roth(2019)]{irr}
Junhwa Hur and Stefan Roth.
\newblock Iterative residual refinement for joint optical flow and occlusion estimation.
\newblock In \emph{Proceedings of the IEEE/CVF Conference on Computer Vision and Pattern Recognition}, pages 5754--5763, 2019.

\bibitem[Ilg et~al.(2017)Ilg, Mayer, Saikia, Keuper, Dosovitskiy, and Brox]{flownet2}
Eddy Ilg, Nikolaus Mayer, Tonmoy Saikia, Margret Keuper, Alexey Dosovitskiy, and Thomas Brox.
\newblock Flownet 2.0: Evolution of optical flow estimation with deep networks.
\newblock In \emph{Proceedings of the IEEE Conference on Computer Vision and Pattern Recognition (CVPR)}, 2017.

\bibitem[Jiang et~al.(2021)Jiang, Campbell, Lu, Li, and Hartley]{gma}
Shihao Jiang, Dylan Campbell, Yao Lu, Hongdong Li, and Richard Hartley.
\newblock Learning to estimate hidden motions with global motion aggregation.
\newblock In \emph{Proceedings of the IEEE/CVF International Conference on Computer Vision}, pages 9772--9781, 2021.

\bibitem[Jiao et~al.(2021)Jiao, Shi, and Tran]{mfrflow}
Yang Jiao, Guangming Shi, and Trac~D Tran.
\newblock Optical flow estimation via motion feature recovery.
\newblock In \emph{2021 IEEE International Conference on Image Processing (ICIP)}, pages 2558--2562. IEEE, 2021.

\bibitem[Kale et~al.(2015)Kale, Pawar, and Dhulekar]{kale2015moving}
Kiran Kale, Sushant Pawar, and Pravin Dhulekar.
\newblock Moving object tracking using optical flow and motion vector estimation.
\newblock In \emph{2015 4th international conference on reliability, infocom technologies and optimization (ICRITO)(trends and future directions)}, pages 1--6. IEEE, 2015.

\bibitem[Kondermann et~al.(2016)Kondermann, Nair, Honauer, Krispin, Andrulis, Brock, Gussefeld, Rahimimoghaddam, Hofmann, Brenner, et~al.]{hd1k}
Daniel Kondermann, Rahul Nair, Katrin Honauer, Karsten Krispin, Jonas Andrulis, Alexander Brock, Burkhard Gussefeld, Mohsen Rahimimoghaddam, Sabine Hofmann, Claus Brenner, et~al.
\newblock The hci benchmark suite: Stereo and flow ground truth with uncertainties for urban autonomous driving.
\newblock In \emph{Proceedings of the IEEE Conference on Computer Vision and Pattern Recognition Workshops}, pages 19--28, 2016.

\bibitem[Li et~al.(2021)Li, Li, and Lu]{dcvc}
Jiahao Li, Bin Li, and Yan Lu.
\newblock Deep contextual video compression.
\newblock \emph{Advances in Neural Information Processing Systems}, 34:\penalty0 18114--18125, 2021.

\bibitem[Loshchilov and Hutter(2018)]{adamw}
Ilya Loshchilov and Frank Hutter.
\newblock Decoupled weight decay regularization.
\newblock In \emph{International Conference on Learning Representations}, 2018.

\bibitem[Lu et~al.(2019)Lu, Ouyang, Xu, Zhang, Cai, and Gao]{dvc}
Guo Lu, Wanli Ouyang, Dong Xu, Xiaoyun Zhang, Chunlei Cai, and Zhiyong Gao.
\newblock Dvc: An end-to-end deep video compression framework.
\newblock In \emph{Proceedings of the IEEE/CVF Conference on Computer Vision and Pattern Recognition}, pages 11006--11015, 2019.

\bibitem[Lu et~al.(2023)Lu, Wang, Ma, Geng, Chen, Chen, and Liu]{transflow}
Yawen Lu, Qifan Wang, Siqi Ma, Tong Geng, Yingjie~Victor Chen, Huaijin Chen, and Dongfang Liu.
\newblock Transflow: Transformer as flow learner.
\newblock In \emph{Proceedings of the IEEE/CVF Conference on Computer Vision and Pattern Recognition}, pages 18063--18073, 2023.

\bibitem[Luo et~al.(2022)Luo, Yang, Luo, Li, Fan, and Liu]{agflow}
Ao Luo, Fan Yang, Kunming Luo, Xin Li, Haoqiang Fan, and Shuaicheng Liu.
\newblock Learning optical flow with adaptive graph reasoning.
\newblock \emph{arXiv preprint arXiv:2202.03857}, 2022.

\bibitem[Luo et~al.(2023)Luo, Yang, Li, Nie, Lin, Fan, and Liu]{gaflow}
Ao Luo, Fan Yang, Xin Li, Lang Nie, Chunyu Lin, Haoqiang Fan, and Shuaicheng Liu.
\newblock Gaflow: Incorporating gaussian attention into optical flow.
\newblock In \emph{Proceedings of the IEEE/CVF International Conference on Computer Vision}, pages 9642--9651, 2023.

\bibitem[Mayer et~al.(2016)Mayer, Ilg, H{\"a}usser, Fischer, Cremers, Dosovitskiy, and Brox]{flyingthings}
N. Mayer, E. Ilg, P. H{\"a}usser, P. Fischer, D. Cremers, A. Dosovitskiy, and T. Brox.
\newblock A large dataset to train convolutional networks for disparity, optical flow, and scene flow estimation.
\newblock In \emph{IEEE International Conference on Computer Vision and Pattern Recognition (CVPR)}, 2016.
\newblock arXiv:1512.02134.

\bibitem[Menze et~al.(2015)Menze, Heipke, and Geiger]{kitti15}
Moritz Menze, Christian Heipke, and Andreas Geiger.
\newblock Joint 3d estimation of vehicles and scene flow.
\newblock In \emph{ISPRS Workshop on Image Sequence Analysis (ISA)}, 2015.

\bibitem[Neoral et~al.(2019)Neoral, {\v{S}}ochman, and Matas]{continualflow}
Michal Neoral, Jan {\v{S}}ochman, and Ji{\v{r}}{\'\i} Matas.
\newblock Continual occlusion and optical flow estimation.
\newblock In \emph{Computer Vision--ACCV 2018: 14th Asian Conference on Computer Vision, Perth, Australia, December 2--6, 2018, Revised Selected Papers, Part IV 14}, pages 159--174. Springer, 2019.

\bibitem[Paszke et~al.(2017)Paszke, Gross, Chintala, Chanan, Yang, DeVito, Lin, Desmaison, Antiga, and Lerer]{pytorch}
Adam Paszke, Sam Gross, Soumith Chintala, Gregory Chanan, Edward Yang, Zachary DeVito, Zeming Lin, Alban Desmaison, Luca Antiga, and Adam Lerer.
\newblock Automatic differentiation in pytorch.
\newblock 2017.

\bibitem[Pont-Tuset et~al.(2017)Pont-Tuset, Perazzi, Caelles, Arbel\'aez, Sorkine-Hornung, and {Van Gool}]{davis17}
Jordi Pont-Tuset, Federico Perazzi, Sergi Caelles, Pablo Arbel\'aez, Alexander Sorkine-Hornung, and Luc {Van Gool}.
\newblock The 2017 davis challenge on video object segmentation.
\newblock \emph{arXiv:1704.00675}, 2017.

\bibitem[Qiu et~al.(2017)Qiu, Yao, and Mei]{pseudo1dconv}
Zhaofan Qiu, Ting Yao, and Tao Mei.
\newblock Learning spatio-temporal representation with pseudo-3d residual networks.
\newblock In \emph{proceedings of the IEEE International Conference on Computer Vision}, pages 5533--5541, 2017.

\bibitem[Ren et~al.(2019)Ren, Gallo, Sun, Yang, Sudderth, and Kautz]{flowfusion}
Zhile Ren, Orazio Gallo, Deqing Sun, Ming-Hsuan Yang, Erik~B Sudderth, and Jan Kautz.
\newblock A fusion approach for multi-frame optical flow estimation.
\newblock In \emph{2019 IEEE Winter Conference on Applications of Computer Vision (WACV)}, pages 2077--2086. IEEE, 2019.

\bibitem[Shi et~al.(2022)Shi, Zhou, Yang, Yin, and Wang]{shi2022csflow}
Hao Shi, Yifan Zhou, Kailun Yang, Xiaoting Yin, and Kaiwei Wang.
\newblock Csflow: Learning optical flow via cross strip correlation for autonomous driving.
\newblock In \emph{2022 IEEE Intelligent Vehicles Symposium (IV)}, pages 1851--1858. IEEE, 2022.

\bibitem[Shi et~al.(2023{\natexlab{a}})Shi, Huang, Bian, Li, Zhang, Cheung, See, Qin, Dai, and Li]{videoflow}
Xiaoyu Shi, Zhaoyang Huang, Weikang Bian, Dasong Li, Manyuan Zhang, Ka~Chun Cheung, Simon See, Hongwei Qin, Jifeng Dai, and Hongsheng Li.
\newblock Videoflow: Exploiting temporal cues for multi-frame optical flow estimation.
\newblock \emph{arXiv preprint arXiv:2303.08340}, 2023{\natexlab{a}}.

\bibitem[Shi et~al.(2023{\natexlab{b}})Shi, Huang, Li, Zhang, Cheung, See, Qin, Dai, and Li]{flowformer++}
Xiaoyu Shi, Zhaoyang Huang, Dasong Li, Manyuan Zhang, Ka~Chun Cheung, Simon See, Hongwei Qin, Jifeng Dai, and Hongsheng Li.
\newblock Flowformer++: Masked cost volume autoencoding for pretraining optical flow estimation.
\newblock In \emph{Proceedings of the IEEE/CVF Conference on Computer Vision and Pattern Recognition}, pages 1599--1610, 2023{\natexlab{b}}.

\bibitem[Smith and Topin(2019)]{onecycle}
Leslie~N Smith and Nicholay Topin.
\newblock Super-convergence: Very fast training of neural networks using large learning rates.
\newblock In \emph{Artificial intelligence and machine learning for multi-domain operations applications}, page 1100612. International Society for Optics and Photonics, 2019.

\bibitem[Sui et~al.(2022)Sui, Li, Geng, Wu, Xu, Liu, Goh, and Zhu]{craft}
Xiuchao Sui, Shaohua Li, Xue Geng, Yan Wu, Xinxing Xu, Yong Liu, Rick Goh, and Hongyuan Zhu.
\newblock Craft: Cross-attentional flow transformer for robust optical flow.
\newblock In \emph{Proceedings of the IEEE/CVF conference on Computer Vision and Pattern Recognition}, pages 17602--17611, 2022.

\bibitem[Sun et~al.(2018)Sun, Yang, Liu, and Kautz]{pwcnet}
Deqing Sun, Xiaodong Yang, Ming-Yu Liu, and Jan Kautz.
\newblock Pwc-net: Cnns for optical flow using pyramid, warping, and cost volume.
\newblock In \emph{Proceedings of the IEEE conference on computer vision and pattern recognition}, pages 8934--8943, 2018.

\bibitem[Sun et~al.(2019)Sun, Yang, Liu, and Kautz]{pwcnet+}
Deqing Sun, Xiaodong Yang, Ming-Yu Liu, and Jan Kautz.
\newblock Models matter, so does training: An empirical study of cnns for optical flow estimation.
\newblock \emph{IEEE transactions on pattern analysis and machine intelligence}, 42\penalty0 (6):\penalty0 1408--1423, 2019.

\bibitem[Sun et~al.(2022)Sun, Chen, Zhu, Guo, and Li]{skflow}
Shangkun Sun, Yuanqi Chen, Yu Zhu, Guodong Guo, and Ge Li.
\newblock Skflow: Learning optical flow with super kernels.
\newblock \emph{Advances in Neural Information Processing Systems}, 35:\penalty0 11313--11326, 2022.

\bibitem[Teed and Deng(2020)]{raft}
Zachary Teed and Jia Deng.
\newblock Raft: Recurrent all-pairs field transforms for optical flow.
\newblock In \emph{European conference on computer vision}, pages 402--419. Springer, 2020.

\bibitem[Wang et~al.(2023)Wang, Zhang, Li, Yu, Sun, Liu, and Hu]{splatflow}
Bo Wang, Yifan Zhang, Jian Li, Yang Yu, Zhenping Sun, Li Liu, and Dewen Hu.
\newblock Splatflow: Learning multi-frame optical flow via splatting.
\newblock \emph{arXiv preprint arXiv:2306.08887}, 2023.

\bibitem[Xu et~al.(2022)Xu, Zhang, Cai, Rezatofighi, and Tao]{gmflow}
Haofei Xu, Jing Zhang, Jianfei Cai, Hamid Rezatofighi, and Dacheng Tao.
\newblock Gmflow: Learning optical flow via global matching.
\newblock In \emph{Proceedings of the IEEE/CVF conference on computer vision and pattern recognition}, pages 8121--8130, 2022.

\bibitem[Yang and Ramanan(2019)]{vcn}
Gengshan Yang and Deva Ramanan.
\newblock Volumetric correspondence networks for optical flow.
\newblock \emph{Advances in neural information processing systems}, 32, 2019.

\bibitem[Yin et~al.(2019)Yin, Darrell, and Yu]{hd3}
Zhichao Yin, Trevor Darrell, and Fisher Yu.
\newblock Hierarchical discrete distribution decomposition for match density estimation.
\newblock In \emph{Proceedings of the IEEE/CVF Conference on Computer Vision and Pattern Recognition}, pages 6044--6053, 2019.

\bibitem[Zhang et~al.(2021)Zhang, Woodford, Prisacariu, and Torr]{sepflow}
Feihu Zhang, Oliver~J Woodford, Victor~Adrian Prisacariu, and Philip~HS Torr.
\newblock Separable flow: Learning motion cost volumes for optical flow estimation.
\newblock In \emph{Proceedings of the IEEE/CVF International Conference on Computer Vision}, pages 10807--10817, 2021.

\bibitem[Zhao et~al.(2020)Zhao, Sheng, Dong, Chang, Xu, et~al.]{maskflownet}
Shengyu Zhao, Yilun Sheng, Yue Dong, Eric~I Chang, Yan Xu, et~al.
\newblock Maskflownet: Asymmetric feature matching with learnable occlusion mask.
\newblock In \emph{Proceedings of the IEEE/CVF Conference on Computer Vision and Pattern Recognition}, pages 6278--6287, 2020.

\bibitem[Zhao et~al.(2022)Zhao, Zhao, Zhang, Zhou, and Metaxas]{gmflownet}
Shiyu Zhao, Long Zhao, Zhixing Zhang, Enyu Zhou, and Dimitris Metaxas.
\newblock Global matching with overlapping attention for optical flow estimation.
\newblock In \emph{Proceedings of the IEEE/CVF Conference on Computer Vision and Pattern Recognition}, pages 17592--17601, 2022.

\end{thebibliography}
}


\end{document}